\documentclass{article}

 \usepackage[preprint]{neurips_2025}

\usepackage[utf8]{inputenc} 
\usepackage[T1]{fontenc}    
\usepackage{hyperref}       
\usepackage{url}            
\usepackage{booktabs}       
\usepackage{amsfonts}       
\usepackage{nicefrac}       
\usepackage{microtype}      
\usepackage{xcolor}         
\usepackage{amsmath}
\usepackage{enumitem}
\usepackage{graphicx}
\usepackage{tabularx}
\usepackage{subcaption}

\usepackage{listings}
\lstnewenvironment{PromptListing}[1][]{%
  \lstset{
    basicstyle=\ttfamily\small,
    breaklines=true,
    breakatwhitespace=true,
    breakautoindent=false,
    breakindent=0pt,
    columns=fullflexible,
    captionpos=t,
    #1
  }
}{}

\usepackage{amsthm}

\newcommand{\repoURL}{\url{https://github.com/kacperkadziolka/causal-reasoning-in-pieces}}

\title{Causal Reasoning in Pieces: Modular In-Context Learning for Causal Discovery}

\author{%
  Kacper Kadziolka and Saber Salehkaleybar\thanks{Corresponding Author.} \\
  Leiden Institute of Advanced Computer Science (LIACS)\\ 
  \texttt{k.r.kadziolka@umail.leidenuniv.nl, s.salehkaleybar@liacs.leidenuniv.nl} \\
}

\begin{document}

\maketitle

\begin{abstract}
    Causal inference remains a fundamental challenge for large language models. Recent advances in internal reasoning with large language models have sparked interest in whether state-of-the-art reasoning models can robustly perform causal discovery—a task where conventional models often suffer from severe overfitting and near-random performance under data perturbations.
    We study causal discovery on the \textsc{Corr2Cause} benchmark using the emergent OpenAI's \textit{o-series} and \textit{DeepSeek-R} model families and find that these reasoning-first architectures achieve significantly greater native gains than prior approaches. To capitalize on these strengths, we introduce a modular in-context pipeline inspired by the Tree-of-Thoughts and Chain-of-Thoughts methodologies, yielding nearly three-fold improvements over conventional baselines. We further probe the pipeline’s impact by analyzing reasoning chain length, complexity, and conducting qualitative and quantitative comparisons between conventional and reasoning models. Our findings suggest that while advanced reasoning models represent a substantial leap forward, carefully structured in-context frameworks are essential to maximize their capabilities and offer a generalizable blueprint for causal discovery across diverse domains.
\end{abstract}

\section{Introduction}

Causal inference—the ability to infer direct cause-effect relationships from observational data underpins progress across the sciences, from epidemiology to economics. While large language models (LLMs) have recently demonstrated extraordinary performance in a variety of reasoning tasks \citep{brown2020languagemodelsfewshotlearners, wei2023chainofthoughtpromptingelicitsreasoning}, their capacity for \emph{robust} causal discovery remains underdeveloped. Evaluations on \textsc{Corr2Cause} reveal that even fine-tuned models collapse under minimal perturbations \citep{jin2024largelanguagemodelsinfer} and often resort to superficial pattern matching rather than genuine causal inference \citep{zečević2023causalparrotslargelanguage}. This exposes a crucial gap between prompt-level heuristics and true causal reasoning.

The \textsc{Corr2Cause} benchmark\footnote{\url{https://huggingface.co/datasets/causal-nlp/corr2cause}} \citep{jin2024largelanguagemodelsinfer} provides a controlled evaluation framework in which each sample encodes statistical correlations and conditional independencies derived from synthetic DAGs. Although fine‐tuned classifiers achieve high in‐distribution accuracy, their performance deteriorates sharply under subtle adversarial perturbations, such as paraphrasing or variable‐identifier substitutions, while in‐context and chain‐of‐thought prompting only yields marginal, non‐robust improvement.

In this work, we reassess \textsc{Corr2Cause} using two state‐of‐the‐art reasoning‐specialist LLM families—OpenAI’s \textit{o3-mini} and \textit{DeepSeek-R1}—which exhibit internally disciplined inference processes via reinforcement-learning self-evolution \citep{deepseekai2025deepseekr1incentivizingreasoningcapability}. By embedding a well-known causal discovery algorithm (Peter–Clark (PC) algorithm, \citep{spirtes2001causation}) within a single, unified prompt, we establish a zero-shot baseline that already surpasses all previously reported results.

\begin{figure}
  \centering
  \includegraphics[width=0.6\columnwidth]{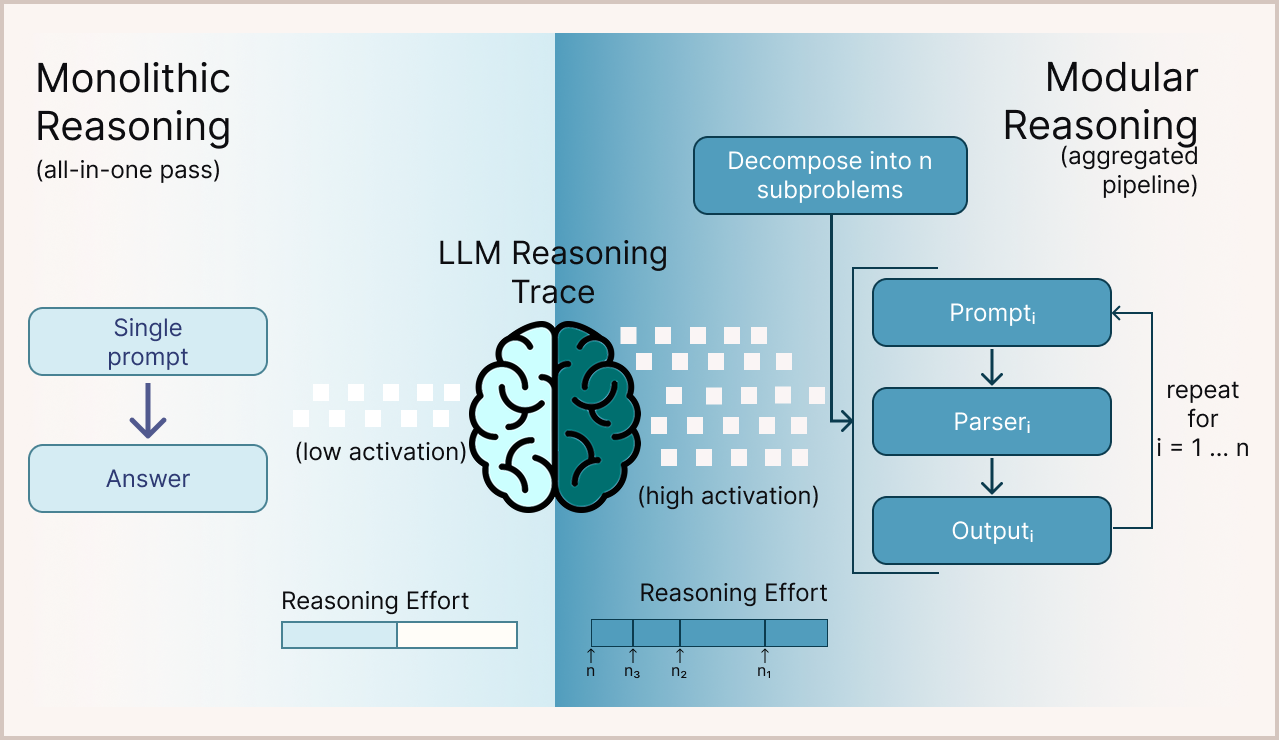}
  \caption{(Left) A single-prompt pass activates the model’s internal reasoning once—shown as low activation, and consumes relatively few tokens (light bar denotes average token usage as reasoning effort). (Right) The modular pipeline splits the task into $n$ prompt–parser–output stages; each stage re-activates the reasoning trace, yielding a cumulatively high activation and substantially greater total token usage (dark segmented bar). This richer, multi-stage reasoning increases overall reasoning effort and produces more robust answers than monolithic baseline.}
  \label{fig:framework_illustration}
\end{figure}

To further improve the performance, we propose a \emph{modular in‐context pipeline framework} tailored to reasoning‐specialist LLMs. Rather than encoding all logic into a single monolithic prompt (Figure \ref{fig:framework_illustration}), our framework guides the model through a user‐defined sequence of \(n\) subproblems, each issued as its own prompt and followed by a lightweight parser. Decomposing the reasoning task in this way yields progressively richer intermediate artifacts, sharpens the model’s focus, and—by aggregating multiple sub‐answers—transforms a brittle one‐shot pass into a robust, multi‐stage inference procedure that substantially improves accuracy.

In summary, we have developed a modular in‐context pipeline tailored for PC‐algorithm causal discovery as evaluated on the \textsc{Corr2Cause} benchmark. By decomposing the task into four focused prompts—undirected skeleton extraction, v‐structure identification, Meek‐rules orientation, and hypothesis evaluation—we achieve up to a three‐fold F1 improvement without fine‐tuning. While this design is optimized for causal‐discovery tasks, exploring its adaptation to other structured‐inference domains represents a promising avenue for future work. All code, prompt templates, and evaluation scripts are available at our GitHub repository.\footnote{\repoURL}

\section{Preliminaries}

This section is dedicated to discuss the \textsc{Corr2Cause} benchmark \citep{jin2024largelanguagemodelsinfer}, which has been designed for evaluating causal discovery capabilities of generative models \citep{bagheri2024textc2textpfeaturinglargelanguage, liu2024llmscapabledatabasedstatistical, wan2025largelanguagemodelscausal}. In what follows, we detail the theoretical foundations of causal discovery assessed in this benchmark dataset, construction methodology, and statistical properties of the dataset. Furthermore, we critically examine the experimental evaluation performed in original work, highlighting limitations associated with fine-tuning and the reliance on chain-of-though (CoT) prompting to our research work. The discussion lays the necessary background and justification for the research questions and methodological innovations presented in subsequent sections.

\subsection{Theoretical Foundations}
\label{subsec:theoretical_foundations}

The \textsc{Corr2Cause} benchmark is grounded in causal discovery using Directed Graphical Causal Models (DGCMs). In this modeling, a causal relationships between $N$ variables are presented by a directed acyclic graph (DAG), where nodes denote variables and there is an edge from $X_i$ to $X_j$ if $X_i$ is the direct cause of $X_j$ (e.g., $ X_i \rightarrow X_j $). Key concepts include:

\textbf{Causal Relationships:} The graphical definitions such as parent-child, ancestor-descendant, and special structures like confounders, colliders, and mediators (see Appendix \ref{appendix:causal-defs}) are fundamental to the dataset’s construction. In this benchmark, these relationships form the basis of the hypotheses, which are expressed as natural language questions. The primary objective is to evaluate whether language models can accurately infer these causal relationships, thereby demonstrating their capability to reason about causal structures.

\textbf{D-Separation \citep{1988i}}: d-separation is used to determine conditional independence between variables, by reading off the underlying causal graph (see Appendix \ref{appendix:causal-defs}). A clear understanding of d-separation is crucial, as it defines which variables are conditionally independent, thereby enabling mapping from the underlying causal structure to the observational independencies.

\textbf{Markov Property and Markov Equivalence:} The Markov property states that each variable is conditionally independent of its non-descendants given its parents, allowing the joint probability distribution to be factorized according to the structure of the DAG. Moreover, different DAGs may induce the same set of joint distributions, resulting in the formation of Markov Equivalence Classes (MECs). For a language model to accurately infer the causal hypotheses in this benchmark, it must not only identify direct causal relationships but also recover MECs. This involves identifying all possible DAGs within an equivalence class to ensure that the inferred causal relationships are consistent with the given conditional independencies among the variables.

\subsection{Construction and Task Formulation}

The \textsc{Corr2Cause} \citep{jin2024largelanguagemodelsinfer}, is built through a systematic pipeline. In brief, for a set of $N$ variables, all possible DAGs are generated by allowing edges $X_i \rightarrow X_j$ only when $i < j$, ensuring acyclicity, with redundant graphs removed via isomorphism checks. d-separation sets are then extracted to capture conditional independence relationships, and the unique DAGs are clustered into Markov Equivalence Classes (MECs) under the faithfulness assumption. For each pair of variables, six standardized causal hypotheses (e.g., ``Is-Parent'', ``Is-Child'', etc.) are generated and assigned a binary validity label $v$ based on whether the hypothesized relation holds across all graphs in the corresponding MEC; these conditional independencies and hypotheses are subsequently verbalized into natural language.

Within this benchmark, the core task is to learn a function 
$$
f: (s, h) \mapsto v,
$$
where $s$ describes the conditional independencies among the variables, $h$ posits a causal relation between two specific variables $X_i$ and $X_j$, and $v \in \{0,1\}$ indicates the validity of the proposed relation.

\subsection{Evaluation and Methodological Perspective}

Thanks to systematic construction pipeline, the \textsc{Corr2Cause} benchmark provides a robust framework for evaluating large language models in a controlled settings for causal discovery.

However, the original experimental methodology reveals critical limitations of most language models in inferring causal relationships. Although extensive fine-tuning yields high in-distribution performance (e.g., \textit{RoBERTa-Large MNLI} achieving an F1 score of $94.74\%$), performance degrades markedly under adversarial perturbations—dropping to $55.45\%$ with paraphrasing and from $85.52\%$ to $46.96\%$ for OpenAI's \textit{GPT-3} under variable refactorization, indicating that models tend to overfit to dataset-specific patterns rather than having causal reasoning. While the benchmark paper also explores in-context learning via Chain-of-Thoughts (CoT), we aim to improve upon these techniques by integrating advanced reasoning models that do not rely solely on extensive fine-tuning.

\subsection{Research Objectives and Study Aims}

Building on the \textsc{Corr2Cause} benchmark, our study is motivated by recent breakthroughs in large language models and their emergent embedded reasoning capabilities. Notably, open‐source model such as \textit{DeepSeek-R1} \citep{deepseekai2025deepseekr1incentivizingreasoningcapability} and proprietary OpenAI's \textit{o3‑mini} \citep{openai2025o3mini} exemplify these advances. \textit{DeepSeek-R1}, which leverages reinforcement learning to induce transparent reasoning behaviors without relying on extensive supervised fine-tuning, and OpenAI's \textit{o3‑mini}, designed for cost-effective reasoning with flexible inference settings. These models provide promising approach for achieving robust causal inference. In this context, we aim to consider:

\textbf{In-Context Learning:} While extensive fine-tuning of non-reasoning models on the \textsc{Corr2Cause} benchmark yields high in-distribution performance, these models fail to generalize to perturbations in dataset (such as node relabeling). We seek to explore whether state-of-the-art reasoning LLMs can effectively perform causal discovery tasks using in-context learning, thereby reducing the reliance on heavy fine-tuning.

\textbf{Prompt Strategies:} Given that reasoning-specialist LLMs (e.g., \textit{DeepSeek-R1}) exhibit emergent, RL-driven internal CoT processes, we investigate novel prompt formulations to enhance their base reasoning capabilities. Our goal is to guide these models more effectively in inferring causal relationships.

\textbf{Modular Causal Discovery Framework:} Inspired by the Tree of Thoughts \citep{yao2023treethoughtsdeliberateproblem} and Chain-of-Thoughts \citep{wei2023chainofthoughtpromptingelicitsreasoning} approaches, we propose a streamlined causal discovery pipeline that decomposes the problem into smaller, manageable sub-tasks with intermediate reasoning stages. By guiding the model through detailed per-stage prompts, we aim to foster deeper, more in-depth reasoning and improve overall performance in causal discovery.

\textbf{Reasoning Processes:} We analyze the reasoning traces produced by reasoning model on identical \textsc{Corr2Cause} samples and contrast them with the answers of conventional models, pinpointing the inference steps and patterns that give reasoning models its edge. We further examine whether cases misclassified by the conventional model correspond to longer, more elaborate reasoning chains—to explore how reasoning length relates to accuracy.

\section{Modular In-Context Pipeline Framework}


We build on the Peter–Clark (PC) algorithm \citep{spirtes2001causation} as the mathematical foundation of our core causal‐discovery method and evaluate on the \textsc{Corr2Cause} benchmark. First, we establish a \emph{baseline in‐context learning} approach that prompts models to judge causal hypotheses from the natural‑language conditional independence statements in Section \ref{subsec:baseline}. Second, we introduce a \emph{modular in‑context causal discovery pipeline} that decomposes PC into distinct reasoning stages in Section \ref{subsec:pipeline}. This pipeline improves interpretability and robustness, yielding a fully in‐context causal‐discovery workflow without any fine‐tuning.

\subsection{Peter-Clark Algorithm}
\label{subsect:pc_algorithm_assumptions}

The PC algorithm \citep{spirtes2001causation} is a well-known constraint‐based causal discovery algorithm, using conditional independence testing to recover a causal graph from purely observational data. It proceeds in two main phases: first, it constructs an undirected skeleton by removing any edge between a pair of variables whenever a conditional independence test succeeds; second, it partially orients the remaining edges. Crucially, the correctness of these steps rests on three assumptions, which together guarantee that the set of conditional independence observed in the data reflects the true underlying causal structure. We now briefly review these assumptions before guiding an LLM through each phase of the algorithm:

\textbf{Directed Acyclic Graphs (DAGs):} It is assumed that the causal relationships among the variables $\mathcal{X} = \{X_1, \dots, X_N\}$ is represented by a DAG $G = (X, E)$, where each directed edge $X_i \to X_j$ denotes a direct causal effect and no directed cycles exist.

\textbf{Causal Markov Condition:} Under the Causal Markov assumption \citep{scheines2005introduction}, each variable $X_i$ in a DAG is independent of its non‑descendants given its direct parents:
$$
X_i \perp\!\!\!\perp \mathrm{NonDesc}(X_i)\;\big|\;\mathrm{Pa}(X_i).
$$
Equivalently, the joint distribution factorizes as $P(X)=\prod_i P(X_i \mid \mathrm{Pa}(X_i))$. For instance, in the chain $A \to B \to C$, one obtains $A \perp\!\!\!\perp C \mid B$. For a distribution generated by a structural causal model.

\textbf{Faithfulness:} Under the faithfulness assumption, the observed distribution exhibits exactly the conditional independence relations implied by the DAG via d‑separation (see Appendix \ref{appendix:causal-defs} for a formal definition)  \citep{10.1214/12-AOS1080}. Equivalently, no conditional independencies exist beyond those entailed by the causal Markov property.

\subsection{Stages of PC Algorithm}
\label{subsec:pc_algo}

The PC algorithm can be decomposed into four sequential stages. This decomposition underlies our modular pipeline framework, issuing each stage as a dedicated prompt and can alternatively be merged into a single, comprehensive baseline. By studying each stage separately, we can systematically review and analyze model responses for that stage. The full PC procedure can be decomposed as follows:

\begin{enumerate}[leftmargin=1.2em]
  \item \textbf{Skeleton Estimation:} Construct the skeleton\footnote{The skeleton of a directed graph is an undirected graph resulted by removing the orientations of edges in the directed graph.} of the true underlying graph by checking each pair $(X_i, X_j)$ for finding a conditional independence $X_i \perp\!\!\!\perp X_j |S$ for some subset $S\subseteq \mathcal{X}\backslash \{X_i,X_j\}$, and removing edge $X_i\textrm{--}X_j$ whenever conditional independence is detected.
  
  \item \textbf{V‑structure Identification:} From the skeleton, identify all triples $(X_i, X_k, X_j)$ where $X_i$ and $X_j$ are both adjacent to $X_k$ but not to each other. Orient these as $X_i \to X_k \leftarrow X_j$ whenever $X_k$ does not appear in any separating set $S$ for $(X_i,X_j)$.
  
  \item \textbf{Edge Orientation via Meek’s Rules \citep{meek2013causalinferencecausalexplanation}:} Having identified the undirected skeleton and all colliders, Meek's rules are applied to orient further edges. Each rule matches a specific subgraph in the current Partially Directed Acyclic Graph (PDAG) and orients one further, subject to the constraints of acyclicity and collider preservation. Repeated application of these rules yields a maximally oriented PDAG, also known as the Completed Partially Directed Acyclic Graph (CPDAG), which represents the Markov Equivalence Class (MEC) of DAGs consistent with the observed conditional independencies.
  
  \item \textbf{Hypothesis Evaluation within the Equivalence Class:} Given the resulting CPDAG, determine whether the target causal hypothesis holds in every valid DAG of its Markov Equivalence Class.
\end{enumerate}

These four steps form the backbone of both our \emph{baseline in‑context learning} (where all are embedded in one prompt) and our \emph{modular pipeline} (where each step is a separate prompt). In the next sections, we describe how each approach instantiates these tasks.

\subsection{Single-Prompt Baseline Assessment}
\label{subsec:baseline}

\begin{table}
  \caption{Single-prompt abstract template for PC algorithm based causal hypothesis evaluation in reasoning models: in-context guidance via persona framing and key principles.}
  \label{tab:baseline-prompt}
  \centering
  \begin{tabular}{@{}p{0.20\textwidth}p{0.75\textwidth}@{}}
    \toprule
    \textit{Preamble}          & Frame the model as a causal‐discovery scientist with PC‐algorithm expertise. \\
    \textit{Task description}  & Evaluate a causal hypothesis using an end‐to‐end, in‐context PC execution. \\
    \textit{Key principles}    & Enumerate the algorithmic steps in skeleton extraction, collider identification, Meek‐rule orientation, and hypothesis validation. \\
    \textit{I/O specification} & Provide a premise (conditional independence statements) and a hypothesis block, and elicit a binary validity judgment. \\
    \bottomrule
  \end{tabular}
\end{table}

To establish a zero-shot performance baseline and probe the native causal-reasoning capabilities of modern reasoning large language models, we employ a single comprehensive prompt that encapsulates all four stages of the PC algorithm described in Section \ref{subsec:pc_algo} within one in-context request. This approach requires minimal prompt engineering, serving both as an accessible native method and as a baseline for a detailed in-context causal discovery pipeline in Section \ref{subsec:pipeline}. By comparing results from this unified prompt against conventional or fine-tuned \textsc{Corr2Cause} models, we can quantify gains attributable to emergent reasoning models without any external fine-tuning.

As detailed in Table \ref{tab:baseline-prompt}, our single‐prompt template comprises four coordinated elements: a persona‐based preamble, a concise task description, key principles, and a unified I/O specification. The persona preamble situates the model in a domain-specific causal-discovery role, an approach shown to improve reasoning consistency and contextual grounding \citep{tseng2024talespersonallmssurvey, tsai2024leveragingllmreasoningenhances}. The task description and key principles explicitly restate the PC algorithm’s procedural steps, skeleton extraction, v-structure identification, applying Meek's rules, and hypothesis validation to steer the model’s internal thinking process. Finally, the I/O specification standardizes a uniform binary output, enabling systematic, fully automated evaluation. This streamlined assessment exploits modern reasoning models’ native strengths with minimal engineering and provides a clear reference point for demonstrating the gains of our modular pipeline \ref{subsec:pipeline}. For full reproducibility, the complete prompt template is provided in Appendix \ref{appendix:baseline}.

\vspace{-0.55cm}
\subsection{Modular In-Context Causal Discovery Pipeline}
\label{subsec:pipeline}

\begin{figure}
  \centering
  \includegraphics[width=0.65\columnwidth]{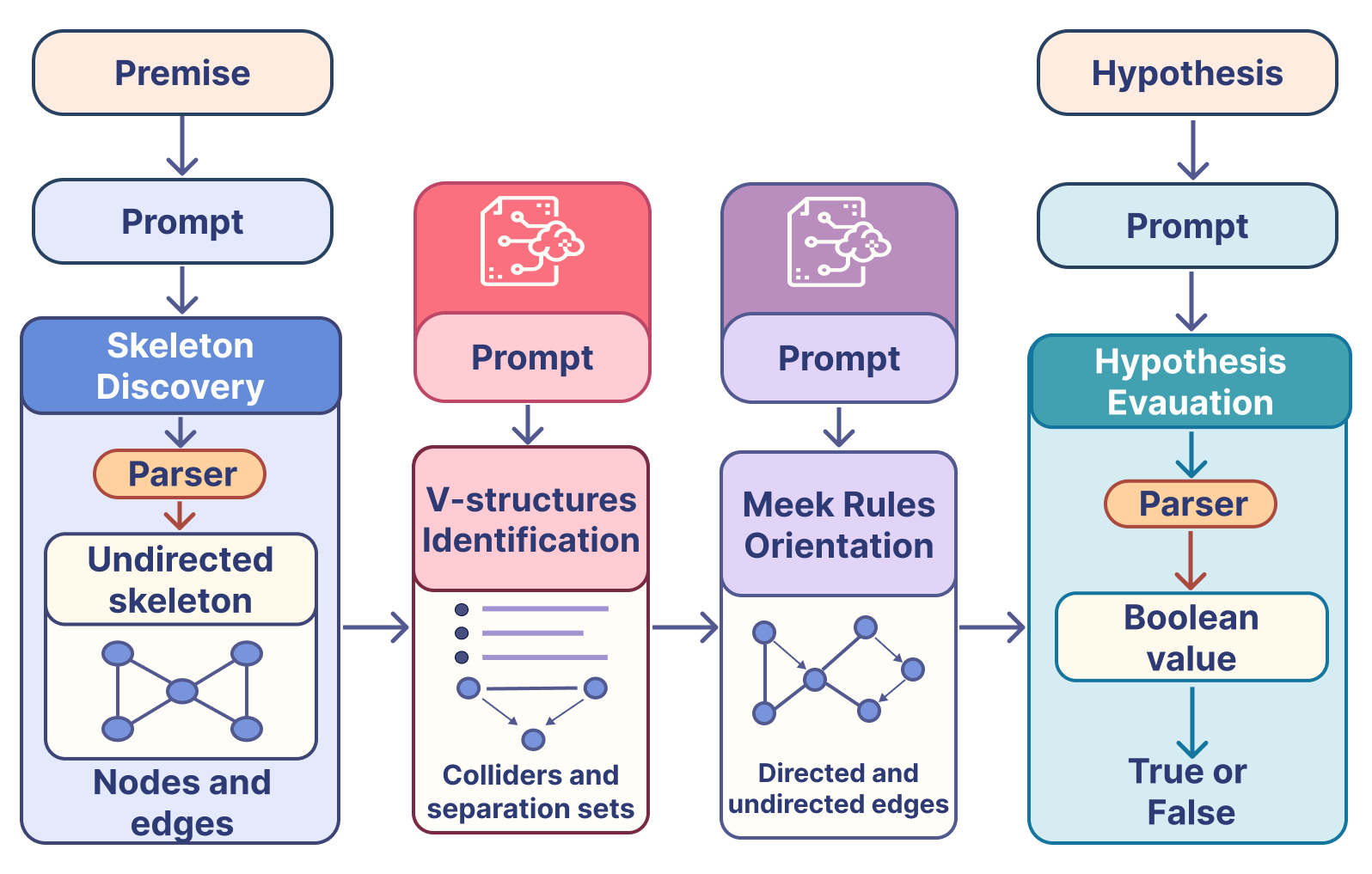}
  \caption{Stage-wise in-context pipeline for causal discovery. Each vertical panel represents a separate prompt and parser corresponding to a stage of the decomposed PC algorithm, with each stage’s output serving as the next stage’s input. Arrows indicate the flow of data between prompts and parsing modules.}
  \label{fig:pipeline_illustration}
\end{figure}

Building upon our zero‐shot baseline, we now introduce a four‐stage in‐context pipeline that decomposes the PC algorithm into separate prompts. This modular framework amplifies the model’s reasoning effort at each stage and produces intermediate artifacts for fine‐grained error analysis, yielding substantial gains over the monolithic baseline on the \textsc{Corr2Cause} benchmark.

Although our single‐prompt approach captures some native reasoning capacity of modern LLMs, it merges four distinct algorithmic operations into one request, making failures difficult to diagnose and limiting the model’s focused attention on each stage. Inspired by Tree‐of‐Thoughts and Chain‐of‐Thoughts methodologies \citep{yao2023treethoughtsdeliberateproblem,wei2023chainofthoughtpromptingelicitsreasoning}, we instead decompose the task into explicit, stage-wise prompts each targeting a single PC stage, and show that this yields both clearer diagnostics and substantial performance gains.

The overall workflow is depicted in Figure \ref{fig:pipeline_illustration}. Each vertical panel comprises an LLM prompt followed by a parser module. Arrows indicate that each stage consumes the previous parser’s output as structured input for the next prompt. By design, no stage requires any external fine‐tuning, only in‐context examples and carefully crafted instructions. Moreover, we standardize across all four prompts the persona framing, input‐schema template, instruction wording, and output‐schema enforcement. Full prompt text appears in Appendix \ref{appendix:pipeline}. Parser modules, implemented in Python, convert raw LLM outputs into canonical Python objects, lists of v-structure triples and edges for onward prompting.

These design choices also affect prompt length and token budget. Although shorter, focused prompts fit more comfortably within model context windows, four sequential calls incur additional latency. In our experiments, the performance gains far outweigh the modest increase in token usage. Our implementation additionally includes automatic re-prompting on schema violations to ensure robustness during empirical evaluation. In sum, this modular in‐context framework unlocks the full potential of reasoning-specialist LLMs and offers a blueprint for causal discovery.

\begin{table}
  \centering
  \caption{Comparative performance on the \textsc{Corr2Cause} benchmark, reporting F1 (primary metric), precision, recall, and accuracy for published benchmark results, zero-shot reasoning-model baselines, and the modular pipeline framework. Within each block, the best value in each column is underlined; the overall best F1 across all methods is both bold and underlined.}
  \label{tab:results}
  \begin{tabular}{lcccc}
    \toprule
    \textbf{Method} & \textbf{F1} & \textbf{Precision} & \textbf{Recall} & \textbf{Accuracy} \\
    \midrule
    \multicolumn{5}{l}{\textbf{\emph{Benchmark results}}} \\
    Random (uniform)  & 20.38             & 15.11             & 31.29             &             62.78 \\
    BART MNLI         & \underline{33.38} & \underline{31.59} & 35.38             & \underline{78.50} \\
    Alpaca-7B         & 27.37             & 15.93             & \underline{97.37} &             21.33 \\
    GPT-4             & 29.08             & 20.92             & 47.66             &             64.60 \\
    \midrule
    \multicolumn{5}{l}{\textbf{\emph{Baseline results}}} \\
    DeepSeek-R1-70B (temp=1)   & 44.93             & 30.39             & 86.11             & 67.24             \\
    DeepSeek-R1-70B (temp=0.6) & 47.02             & 32.54             & 84.75             & 70.09             \\
    DeepSeek-R1-70B (temp=0.1) & 48.56             & 33.76             & \underline{86.49} & 70.80             \\
    DeepSeek-R1 API (temp=0.1) & 64.57             & 53.33             & 81.82             & 86.02             \\
    OpenAI o3-mini             & \underline{66.28} & \underline{70.19} & 62.78             & \underline{90.10} \\
    \midrule
    \multicolumn{5}{l}{\textbf{\emph{Pipeline framework results}}} \\
    DeepSeek-R1-70B (temp=0.1) &             65.33          & 55.88             & 78.62             & 87.28             \\
    DeepSeek-R1 API (temp=0.1) &             79.83          & 73.40             & \underline{87.65} & 93.27             \\
    OpenAI o3-mini             & \underline{\textbf{83.83}} & \underline{90.91} & 77.78             & \underline{95.32} \\
    \bottomrule
  \end{tabular}
\end{table}
\subsection{Comparison with the Related Work}
There is a very recent work called PC-SubQ \citep{sgouritsa2024prompting}, which also adopted a staged prompting strategy for causal discovery with PC algorithm. In comparison with this work, there are substantial differences in both methodology and outcomes:1- Different modular design: PC-SubQ decomposes the PC algorithm into eight sub-steps, whereas our framework consists of four broader modules: skeleton discovery, v-structure identification, Meek rule application, and hypothesis evaluation. In our framework, each module is paired with custom prompt templates and Python-based parser modules to structure and pass intermediate outputs across stages. Apparently, this part is not performed in PC-SubQ.
2- Quantitative performance gap: Our modular pipeline achieves an F1 score of $83.83\%$ on CORR2CAUSE with OpenAI o3-mini—a nearly $18\%$ improvement over the baseline reasoning model ($66.28\%$). In contrast, PC-SubQ reports substantially lower F1 scores, which are less than $50\%$ in most considered models. While some differences may stem from model choices, we note that PC-SubQ reports less than $10\%$ improvement over the best considered baselines across nearly all models they evaluated. 3- Reasoning model focus and analysis: Our work focuses specifically on reasoning-centric architectures (OpenAI o3-mini, DeepSeek-R1), and in Section \ref{sec:qualitative-quantitative-analysis}, we provide detailed quantitative and qualitative analyses to explain why these models perform better. Our pipeline further amplifies this advantage, showing that even strong reasoning models benefit significantly from modular in-context design.

\vspace{-0.3cm}
\section{Empirical Evaluation and Analysis}


We present a threefold evaluation of our causal‐discovery methods. In Section \ref{sec:benchmark-comparison}, we benchmark our zero‐shot reasoning baselines and modular in-context pipeline against published \textsc{Corr2Cause} results. Section \ref{sec:intermodel-comparison} offers an inter‐model analysis of reasoning LLMs, identifying which pipeline stages are most error‐prone. Finally, Section \ref{sec:qualitative-quantitative-analysis} delivers a combined quantitative and qualitative failure analysis of conventional‐model errors and the reasoning model’s corrective mechanisms. All experiments were conducted under the hardware, software, and model configurations detailed in Appendix \ref{appendix:exp-setup}.

\subsection{Comparative Benchmarking}
\label{sec:benchmark-comparison}

We evaluate on the full test split of the \textsc{Corr2Cause} benchmark, reporting F1 (primary metric), precision, recall, and accuracy for published baselines, our zero‐shot reasoning models, and the four‐stage pipeline, Table \ref{tab:results}. Among the published results, \textit{BART MNLI} achieves the highest F1 ($33.38$), reflecting the limitations of conventional language models. In contrast, zero‐shot reasoning models deliver dramatic native improvements. The self‐hosted \textit{DeepSeek-R1} (70B parameters) attains an F1 of $48.56$, while the much larger \textit{DeepSeek-R1 API} (685B parameters) and OpenAI's \textit{o3-mini} (size undisclosed) reach F1 of $64.57$ and $66.28$, respectively. These gains show both the power of emergent reasoning models and the impact of model scale when applied without any additional fine‐tuning.

Proposed modular pipeline further amplifies performance, \textit{DeepSeek-R1 API}’s F1 jumps to $79.83$, and the OpenAI's \textit{o3-mini} pipeline achieves $83.83$. Notably, these improvements are accompanied by more balanced precision and recall despite the dataset’s approximately $85$\% "false" label prevalence. Recall increases by roughly $10$–$15$ percentage points over the zero‐shot baselines, accompanied by even larger gains in precision, indicating that our stage‐wise prompts help the model avoid a trivial "always-false" strategy. Together, these results validate that (i) reasoning‐specialist LLMs substantially outperform prior models in a pure zero‐shot setting and (ii) a carefully structured, in‐context pipeline can unlock further gains, achieved solely through prompt design and stage decomposition, without external supervision or fine‐tuning.

\subsection{Inter‐Model Comparison}
\label{sec:intermodel-comparison}

\begin{figure}
  \centering
  \includegraphics[width=0.9\columnwidth]{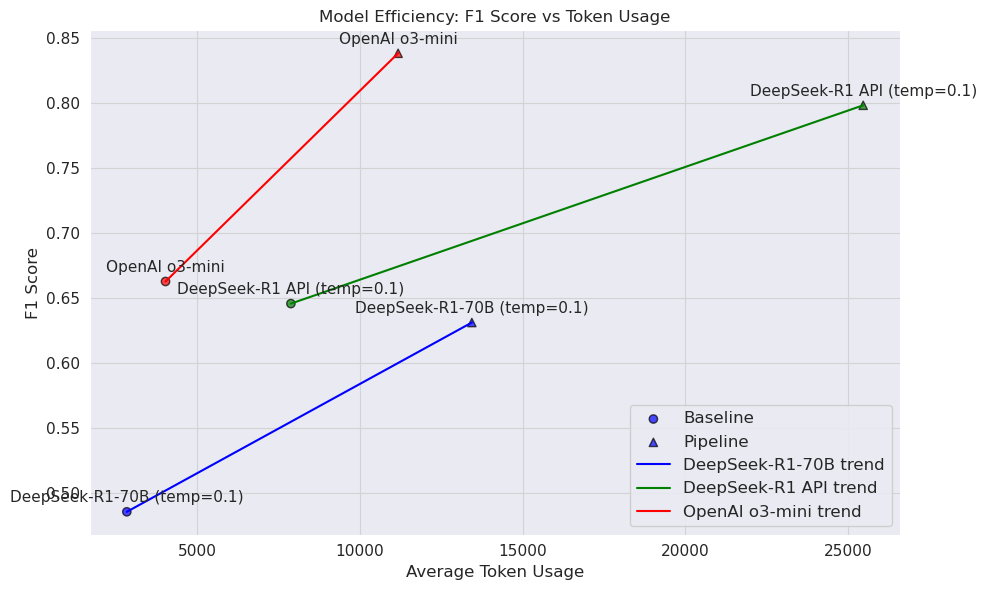}
  \caption{Efficiency trade-off on the \textsc{Corr2Cause} benchmark. Scatter points denote F1 scores against average token usage for both baseline (circles) and pipeline (triangles) settings. Solid lines connect each model’s baseline and pipeline points as trend lines, highlighting how increased token investment in the pipeline correlates with F1 improvements.}
  \label{fig:model_efficiency}
\end{figure}

Figure \ref{fig:model_efficiency} plots each model’s zero-shot baseline and full-pipeline F1 scores against its average token usage. Immediately evident is the linear uplift that our modular pipeline yields over the same monolithic prompt—every triangle sits above its corresponding circle, demonstrating that decomposing the PC algorithm into stage-wise subprompts delivers consistent gains without any additional fine-tuning. Under the identical single-prompt setup, the \textit{DeepSeek-R1 API} (685B parameters) outperforms its 70B counterpart by approximately 15 $F1$ points, underscoring our hypothesis that \emph{more intrinsic reasoning capacity} leads to better causal‐discovery performance.

OpenAI’s \textit{o3-mini} not only achieves the highest end-to-end F1 but also the most favorable trade-off between token cost and accuracy—rising from roughly $4K$ to $11K$ tokens for a $+0.17$ points gain in F1 , versus \textit{DeepSeek-R1 API}'s $8K \rightarrow 25K$ tokens for a comparable improvement. Thus, while larger models can invest more computation, the \textit{o3-mini} attains superior performance per token when orchestrated through our pipeline.

\begin{table}
  \centering
  \caption{Stage-wise F1 scores for each pipeline tested reasoning model.}
  \label{tab:stage_wise_f1}
  \begin{tabular}{lrrrr}
    \toprule
    \textbf{Model} & \textbf{Skeleton} & \textbf{V-structures} & \textbf{Meek’s Rules} & \textbf{Hypothesis} \\
    \midrule
    DeepSeek-R1-70B & 0.83 & 0.76 & 0.70 & 0.58 \\
    DeepSeek-R1 API & 1.00 & 0.99 & 0.90 & 0.80 \\
    OpenAI o3-mini  & 1.00 & 0.99 & 0.95 & 0.84 \\
    \bottomrule
  \end{tabular}
\end{table}

Table \ref{tab:stage_wise_f1} summarizes each model’s F1 at the four pipeline stages. All three models achieve near‐perfect skeleton discovery and v-structure identification steps, but diverge later under Meek’s rules and hypothesis evaluation phases, with the \textit{o3-mini} retaining the strongest performance. Taken together, these comparisons show (i) our modular pipeline uniformly boosts every model’s F1 via focused, stage-wise reasoning; (ii) larger LLMs begin already perform better on the initial stages; and (iii) OpenAI \textit{o3-mini} achieves the best overall trade-off of token cost and final accuracy.

\subsection{Qualitative and Quantitative Failure Analysis}
\label{sec:qualitative-quantitative-analysis}

\begin{figure}
  \centering
  \includegraphics[width=0.9\columnwidth]{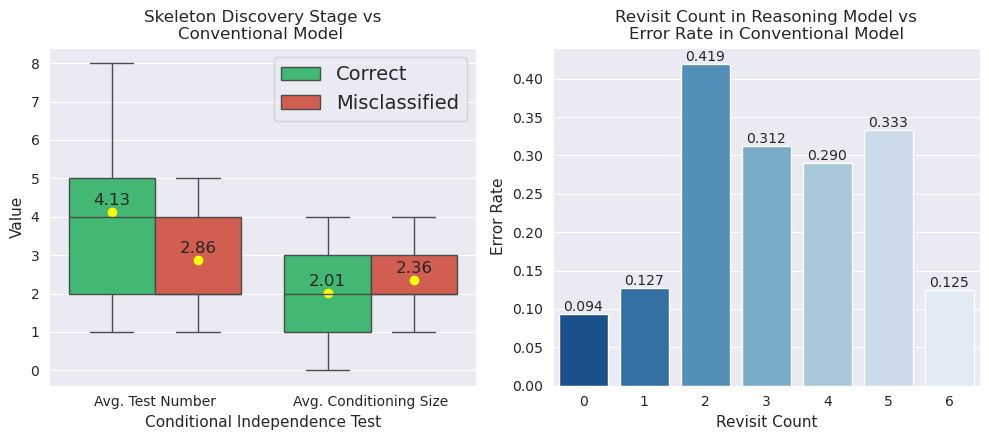}
  \caption{(Left) Boxplot of conditional independence sets metrics—number of tests and maximum conditioning size reported for the conventional model. Edges are labeled “Correctly classified” if the skeleton matches the ground truth, and “Misclassified” if it includes incorrect edges or leaves out true ones. Yellow dots indicate mean values. (Right) Bars show, for each revisit count $k$, the fraction of edges that the common model misclassified among those the reasoning model revisited $k$ times.}
  \label{fig:quant_qual_analysis}
\end{figure}


We use \textit{LLaMA3.3-70B} as the conventional model and \textit{DeepSeek-R1-70B} (distilled from \textit{LLaMA3.3-70B}) as the reasoning model. We selected approximately $100$ first stage samples where the common model made at least one error; the reasoning model attained perfect accuracy on these. Behaviorally, the reasoning model performs iterative self‐checks (“Wait…”, “Hold on…”), explicitly revisits each edge during reasoning trace micro‐steps (mean of $66$ micro‐steps, maximum of $135$ micro‐steps), and refines decisions in real time. While specific patterns vary, this “thinking aloud” and iterative refinement underlie its qualitative advantage. 

Figure \ref{fig:quant_qual_analysis} (left) compares the average number of independence tests of pair of variables that common models classify correctly versus those it misclassifies. Misclassified edges have fewer tests (mean $2.86$ vs $4.13$). It is important to note that the presence of even a single conditional independence between a given pair of variables is sufficient to remove the edge connecting them in the skeleton. The common models often require multiple conditional independence assertions involving the same pair of variables in the premise to remove the edge between them from the skeleton. We also plot the average size of the conditioning set in a conditional independence test (i.e., the size of $Z$ in a test of the form $X \perp\!\!\!\perp Y \mid Z$) for both correctly and misclassified instances. As observed, the average size of the conditioning set tends to be larger in the misclassified class.

The right panel shows the average error of the conventional model on instances where the reasoning model revisits \( k \) times in its reasoning trace.
This result shows frequent re-evaluations as a key mechanism by which the reasoning pipeline corrects the common model’s errors.

\section{Conclusions}
\label{sec:conclusions}


We have shown that state-of-the-art reasoning models surpass prior causal-discovery results on the \textsc{Corr2Cause} benchmark in a pure zero shot baseline settings. Building on this native capability, our four-stage modular in-context pipeline yields up to a three-fold F1 improvement without any fine tuning. Key to these gains are (i) decomposition of the PC algorithm into separate stages; (ii) producing rich intermediate artifacts, and (iii) amplifying the model's native iterative enrichment of its reasoning traces.

These improvements come at the cost of increased computation: multiple sequential model requests and longer reasoning traces substantially raise token usage. Furthermore, our evaluation is confined to a single causal‐discovery benchmark, so assessing robustness and effectiveness on other datasets remains future work.


\bibliographystyle{apalike}
\bibliography{references}

\begin{thebibliography}{}

\bibitem[Bagheri et~al., 2024]{bagheri2024textc2textpfeaturinglargelanguage}
Bagheri, A., Alinejad, M., Bello, K., and Akhondi-Asl, A. (2024).
\newblock $\text{C}^2\text{P}$: Featuring large language models with causal reasoning.

\bibitem[Brown et~al., 2020]{brown2020languagemodelsfewshotlearners}
Brown, T.~B., Mann, B., Ryder, N., Subbiah, M., Kaplan, J., Dhariwal, P., Neelakantan, A., Shyam, P., Sastry, G., Askell, A., Agarwal, S., Herbert-Voss, A., Krueger, G., Henighan, T., Child, R., Ramesh, A., Ziegler, D.~M., Wu, J., Winter, C., Hesse, C., Chen, M., Sigler, E., Litwin, M., Gray, S., Chess, B., Clark, J., Berner, C., McCandlish, S., Radford, A., Sutskever, I., and Amodei, D. (2020).
\newblock Language models are few-shot learners.

\bibitem[DeepSeek-AI, 2025]{deepseekai2025deepseekr1incentivizingreasoningcapability}
DeepSeek-AI (2025).
\newblock Deepseek-r1: Incentivizing reasoning capability in llms via reinforcement learning.

\bibitem[Jin et~al., 2024]{jin2024largelanguagemodelsinfer}
Jin, Z., Liu, J., Lyu, Z., Poff, S., Sachan, M., Mihalcea, R., Diab, M., and Schölkopf, B. (2024).
\newblock Can large language models infer causation from correlation?

\bibitem[Liu et~al., 2024]{liu2024llmscapabledatabasedstatistical}
Liu, X., Wu, Z., Wu, X., Lu, P., Chang, K.-W., and Feng, Y. (2024).
\newblock Are llms capable of data-based statistical and causal reasoning? benchmarking advanced quantitative reasoning with data.

\bibitem[Meek, 2013]{meek2013causalinferencecausalexplanation}
Meek, C. (2013).
\newblock Causal inference and causal explanation with background knowledge.

\bibitem[OpenAI, 2025]{openai2025o3mini}
OpenAI (2025).
\newblock Openai o3-mini: Pushing the frontier of cost-effective reasoning.
\newblock \url{https://openai.com/research/openai-o3-mini}.
\newblock Accessed: April 5, 2025.

\bibitem[Pearl, 1988]{1988i}
Pearl, J. (1988).
\newblock The morgan kaufmann series in representation and reasoning.
\newblock In {\em Probabilistic Reasoning in Intelligent Systems}, page~i. Morgan Kaufmann, San Francisco (CA).

\bibitem[Scheines, 2005]{scheines2005introduction}
Scheines, R. (2005).
\newblock An introduction to causal inference.
\newblock Technical report, Carnegie Mellon University.

\bibitem[Sgouritsa et~al., 2024]{sgouritsa2024prompting}
Sgouritsa, E., Aglietti, V., Teh, Y.~W., Doucet, A., Gretton, A., and Chiappa, S. (2024).
\newblock Prompting strategies for enabling large language models to infer causation from correlation.
\newblock {\em arXiv preprint arXiv:2412.13952}.

\bibitem[Spirtes et~al., 2001]{spirtes2001causation}
Spirtes, P., Glymour, C., Scheines, R., and Heckerman, D. (2001).
\newblock {\em Causation, Prediction, and Search}.
\newblock Adaptive Computation and Machine Learning. The MIT Press, Cambridge, MA, 2 edition.
\newblock Special Collection: CogNet.

\bibitem[Tsai et~al., 2024]{tsai2024leveragingllmreasoningenhances}
Tsai, A.~Y., Kraft, A., Jin, L., Cai, C., Hosseini, A., Xu, T., Zhang, Z., Hong, L., Chi, E.~H., and Yi, X. (2024).
\newblock Leveraging llm reasoning enhances personalized recommender systems.

\bibitem[Tseng et~al., 2024]{tseng2024talespersonallmssurvey}
Tseng, Y.-M., Huang, Y.-C., Hsiao, T.-Y., Chen, W.-L., Huang, C.-W., Meng, Y., and Chen, Y.-N. (2024).
\newblock Two tales of persona in llms: A survey of role-playing and personalization.

\bibitem[Uhler et~al., 2013]{10.1214/12-AOS1080}
Uhler, C., Raskutti, G., B{\"u}hlmann, P., and Yu, B. (2013).
\newblock {Geometry of the faithfulness assumption in causal inference}.
\newblock {\em The Annals of Statistics}, 41(2):436 -- 463.

\bibitem[Wan et~al., 2025]{wan2025largelanguagemodelscausal}
Wan, G., Lu, Y., Wu, Y., Hu, M., and Li, S. (2025).
\newblock Large language models for causal discovery: Current landscape and future directions.

\bibitem[Wei et~al., 2023]{wei2023chainofthoughtpromptingelicitsreasoning}
Wei, J., Wang, X., Schuurmans, D., Bosma, M., Ichter, B., Xia, F., Chi, E., Le, Q., and Zhou, D. (2023).
\newblock Chain-of-thought prompting elicits reasoning in large language models.

\bibitem[Yao et~al., 2023]{yao2023treethoughtsdeliberateproblem}
Yao, S., Yu, D., Zhao, J., Shafran, I., Griffiths, T.~L., Cao, Y., and Narasimhan, K. (2023).
\newblock Tree of thoughts: Deliberate problem solving with large language models.

\bibitem[Zečević et~al., 2023]{zečević2023causalparrotslargelanguage}
Zečević, M., Willig, M., Dhami, D.~S., and Kersting, K. (2023).
\newblock Causal parrots: Large language models may talk causality but are not causal.

\end{thebibliography}


\appendix

\section{Formal Definitions of Causal Concepts}
\label{appendix:causal-defs}

\begin{figure}[!htbp]
  \centering
  \setlength\tabcolsep{8pt}
  \renewcommand{\arraystretch}{1.5}
  \begin{tabular}{@{} m{3cm} m{10cm} @{}}
    \includegraphics[height=2.5cm]{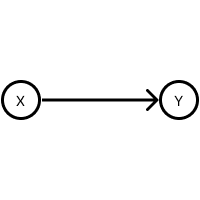}
      & \textbf{Parent–child:} 
        $X$ is a \emph{parent} of $Y$ if there is a directed edge $X\to Y$ in the DAG. \\
    \includegraphics[height=2.5cm]{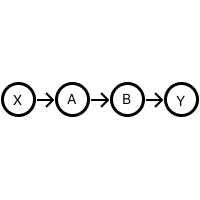}
      & \textbf{Ancestor–descendant:} 
        $X$ is an \emph{ancestor} of $Y$ if there exists a directed path $X\to \dots \to Y$. \\
    \includegraphics[height=2.5cm]{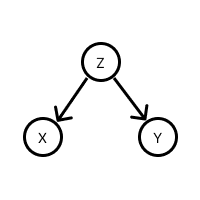}
      & \textbf{Confounder (common cause):} 
        $Z$ is a \emph{confounder} for $X$ and $Y$ if $Z$ is a common parent of both. \\
    \includegraphics[height=2.5cm]{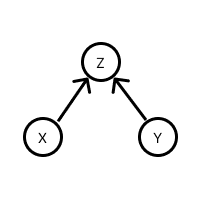}
      & \textbf{Collider (common effect):} 
        Nodes $X$ and $Y$ form a \emph{collider} at $Z$ if $X\to Z\leftarrow Y$ and $X,Y$ are not adjacent. \\
  \end{tabular}
  \caption{Graphical and formal definitions of the four fundamental causal relationships underpinning the 
    \textsc{Corr2Cause} benchmark. These primitives form the basis of each natural-language 
    hypothesis evaluated by our models.}
  \label{fig:causal_relationships}
\end{figure}

In this appendix, we explain the core causal concepts used in both the \textsc{Corr2Cause} benchmark and the field of causal discovery. \textsc{Corr2Cause} turns each question into a binary decision task: it ask whether a certain pattern appears in a causal graph, based on four fundamental relationships that are described in Figure \ref{fig:causal_relationships}. For example, one question might read: \textit{'Hypothesis: There exists at least one collider (i.e., common effect) of C and D.'}.

Conditional independence in Peter-Clark is determined by the appliance of d‐separation criterion. Let $G$ be a DAG and let $X$, $Y$ be two nodes, with $Z$ an arbitrary set of nodes.  A path $\pi$ between $X$ and $Y$ is \emph{blocked by} $Z$ if at least one of the following holds for some consecutive triple $A\!-\!B\!-\!C$ on $\pi$:

\begin{itemize}[nosep]
  \item \textbf{Chain or fork:} There exists $B$ such that $A \to B \to C$ or $A \leftarrow B \to C$, and $B \in Z$.
  \item \textbf{Collider:} There exists $B$ such that $A \to B \leftarrow C$, and neither $B$ nor any descendant of $B$ lies in $Z$.
\end{itemize}

Then $X$ and $Y$ are \emph{d‐separated} by $Z$, written
\[
  X \!\perp\!\!\!\perp_{d}\! Y \mid Z,
\]
if every path between them is blocked by $Z$.  In the PC algorithm, under the causal faithfulness assumption, testing $X \perp\!\!\!\perp Y \mid Z$ implies removal of the edge $X\!-\!Y$ from the skeleton.

\section{Empirical Evaluation Environment}
\label{appendix:exp-setup}

The empirical evaluation was conducted using two access modes. Self-hosted experiments on GPU clusters used the 70B parameter models for both baseline and methodology development, while larger-scale runs leveraged third-party APIs. 

\subsection*{GPU Cluster Settings for Self-Hosted Models}

\textbf{LLMs:} The \emph{LlaMa3.3-70B-Instruct}\footnote{\url{https://huggingface.co/meta-llama/Llama-3.3-70B-Instruct}} version served as our conventional baseline. The \emph{DeepSeek-R1-Distill-Llama-70B}\footnote{\url{https://huggingface.co/deepseek-ai/DeepSeek-R1-Distill-Llama-70B}}—a 70B parameter version of open-source \emph{DeepSeek-R1} distilled from the \emph{LlaMa3.3-70B} was our self-hosted reasoning model.

\textbf{Compute node:} 2× NVIDIA H100-94 GB GPUs, 32 vCPUs, 360 GB RAM.

\textbf{Inference settings:} We perform inference in batches of four examples to maximize GPU utilization. Temperature parameter is listed in Table \ref{tab:results}.

\textbf{Runtime:} Running the zero‐shot baseline over the full $1,162$‐sample test split requires about $47$ GPU‐hours per model. The four‐stage pipeline takes roughly $220$ GPU‐hours on the same split. We ran three independent baseline experiments and one pipeline experiment, for a combined total of about $360$ GPU‐hours. We do not include additional research runs in this tally.

\subsection*{API Hosted Models}

\textbf{LLMs:} We accessed two large reasoning LLMs via paid APIs: 685B version of \emph{DeepSeek-R1}\footnote{\url{https://api-docs.deepseek.com/}} and OpenAI's \emph{o3-mini}\footnote{\url{https://platform.openai.com/docs/models/o3-mini}}.

\textbf{Inference Settings:} All \emph{DeepSeek-R1} calls used the temperature values reported in Table \ref{tab:results}. OpenAI’s \emph{o3-mini} does not expose a temperature parameter, instead it provides a \texttt{reasoning\_effort} control, which we left at its default.

\textbf{Token Usage:} We quantify runtime by total tokens consumed. For the zero-shot baseline, \emph{DeepSeek-R1} used \textasciitilde $8$ k tokens per sample and \emph{o3-mini} \textasciitilde $4$ k tokens. Under the four-stage pipeline, \emph{o3-mini} consumed \textasciitilde $11$ k tokens per sample and \emph{DeepSeek-R1} over $25$ k tokens. Overall, \emph{o3-mini} experiments totaled \textasciitilde $18$ M tokens (input + output), while \emph{DeepSeek-R1} totaled \textasciitilde $53$ M tokens.

\section{Single-Prompt Baseline Template}
\label{appendix:baseline}

We establish a zero-shot baseline by issuing the single-prompt template summarized in Table \ref{tab:baseline-prompt}. The complete prompt text is given in Listing \ref{lst:baseline-prompt}.

\begin{PromptListing}[caption={Single-prompt template for baseline evaluation on the \textsc{Corr2Cause} benchmark.},label={lst:baseline-prompt}]
You are a scientist specializing in causal discovery algorithms, particularly the Peter-Clark (PC) algorithm. You expertly apply correlation statements to initialize graphs and use independence assertions to compute causal undirected skeletons. You have the skill to leverage these skeletons and separation sets to identify v-structures and then translate them into a maximally oriented graph by applying Meek rules. You can also evaluate hypotheses about specific causal relationships between variables, determining whether the provided correlation and independence statements support those hypotheses.
  
You are provided with the following inputs:
- **Premise:** A set of statements containing observed correlations along with both marginal and conditional independence relationships.
- **Hypothesis:** A statement that posits a particular causal relationship between one or more variables. This claim could assert a direct causal link (for example, 'E directly causes A') or an indirect causal relationship (for example, 'B causes something else which causes A'). The hypothesis specifies what causal connection is being proposed and is what you should evaluate based on the given statistical relations and independence statements.
  
**Task:** Use the PC (Peter-Clark) algorithm to assess whether the provided causal claim in the hypothesis is supported by the relationships and independence statements in the premise.
  
**Key Principles:**
- Extract the causal undirected skeleton by interpreting the correlations and conditional independencies.
- Identify v-structures (colliders) that arise from the undirected skeleton and separation sets.
- Apply Meek rules to further orient remaining undirected edges.
- Evaluate the hypothesis based on the processed causal graph.
  
Please reason step by step, and after completing your analysis, provide your final answer only for the hypothesis as a boolean value (either true or false) in this exact JSON format:
  
```json
{{
  "hypothesis_answer": true/false
}}
```

**Inputs:**
Premise: {premise}
Hypothesis: {hypothesis}
\end{PromptListing}

\section{Stage-Wise Pipeline Prompt Templates}
\label{appendix:pipeline}

In this appendix, we provide the exact in‐context prompts used at each stage of our modular PC–algorithm pipeline described in Section \ref{subsec:pipeline}. These prompts were refined through a careful, iterative prompt engineering process: we drafted multiple variants and validated each in a sandbox using programmatic JSON‐schema checks to ensure correct, schema-compliant outputs. Every prompt is accompanied by a dedicated parser module in our codebase to extract structured results.

\subsection*{Stage 1: Undirected Skeleton Extraction}

\begin{PromptListing}[caption={Undirected skeleton discovery stage prompt template.},label={lst:prompt-skeleton}]
You are a scientist specializing in causal discovery algorithms, particularly the Peter-Clark (PC) algorithm. Your expertise lies in using correlation statements and independence assertions to initialize graphs and compute causal undirected skeletons. In this stage, focus exclusively on analyzing the provided correlation and independence information to accurately construct the underlying undirected graph that represents the relationships among the variables.
  
**Task:** Based on the given Premise, apply the PC (Peter-Clark) algorithm to identify a causal undirected skeleton from the correlations and independence statements.

**Key Principles:**
- Initially, assume all correlated variables are connected
- An edge X--Y should be removed if ANY independence statement shows X and Y are independent (marginally or conditionally)
- Keep all edges unless contradicted by an explicit independence statement
- The final skeleton should reflect all independence relationships in the data

Your analysis should be thorough but focused on removing edges based on independence statements.

**Required Output Format:**
  After completing your analysis, provide your final answer in this exact JSON format:

```json
{{
  "nodes": ["list", "of", "nodes"],
  "edges": [
    ["Node1", "Node2"],
    ["Node2", "Node3"]
  ]
}}
```

**Inputs:**
Premise: {premise}
\end{PromptListing}

\subsection*{Stage 2: V-Structure Identification}

\begin{PromptListing}[caption={V-structure identification stage prompt template.},label={lst:prompt-vstructures}]
You are an expert in causal discovery, specializing in the Peter-Clark (PC) algorithm. With a deep understanding of undirected skeletons and separation sets, your task in this stage is to identify v-structures. Analyze the given undirected skeleton along with the corresponding separation sets to pinpoint the colliders that indicate potential directional causal relationships among the variables.
  
**Task:** You are given a causal skeleton (an undirected graph produced by the PC algorithm) and a Premise that contains independence statements. Your job is to perform two distinct steps:

1. **Extraction of Independence Statements:**
  - **Parse the Premise:** Identify and extract all independence statements.
  - **Representation:** For each statement that indicates that a pair of variables is independent (optionally given a conditioning set), represent it as an entry in a dictionary. Use a key that is a sorted pair of variables (e.g., "A,C") and set its value as a list of the conditioning variables (if any).
2. **Identification of V-Structures (Colliders):**
  - **Candidate Identification:** Systematically consider every triple of nodes (X,Z,Y) from the skeleton where:
    - There are edges X--Z and Y--Z in the skeleton.
    - There is no direct edge between X and Y (i.e., they are non-adjacent).
  - **Verification via Separation Test:**
    For each candidate triple, check the separation (independence) information:
      - **Valid V-Structure:** Include [X,Z,Y] only if for the pair (X,Y) (as found in the independence statements) the corresponding separation set does not contain Z.
      - **Systematic Check:** Ensure that every candidate triple is evaluated using the criteria of non-adjacency, common neighbor, and the separation test to avoid false positives (including a Z that appears in the separation set) and false negatives (omitting any valid candidate).

**Required Output Format:**
After completing your analysis, provide your final answer in this exact JSON format:
```json
{{
  "separation_sets": {{
    "A,C": ["B"],
    "A,D": [...],
    ...
  }},
  "v_structures": [
    ["X", "Z", "Y"],
    ["X2", "Z2", "Y2"],
    ...
  ]
}}
```

**Inputs:**
Premise: {premise}
Casual skeleton:
```json
{{
  "nodes": {nodes},
  "edges": {edges}
}}
```
\end{PromptListing}

\subsection*{Stage 3: Meek’s Rules Orientation}

\begin{PromptListing}[caption={Meek's rules orientation stage prompt template.},label={lst:prompt-meekrules}]
You are an expert in causal inference with deep knowledge of the PC algorithm's Meek orientation rules. Your task is to convert a given undirected causal skeleton into a partially directed acyclic graph (PDAG) by orienting as many edges as possible while following these rules:

**Rules and Constraints:**
- **Cycle Avoidance:** Do not create any directed cycles.
- **V-Structure Preservation:** Maintain all given v-structures (i.e., collider configurations).
- **Independence Consistency:** Ensure all orientations comply with the provided marginal and conditional independence relationships (see "Premise").
- **Edge Pool Restriction:** Only orient edges that exist in the provided causal skeleton.
- **Conservative Orientation:** Only assign a direction to an edge if it is uniquely compelled by a v-structure or through propagation via Meek rules. If the orientation is ambiguous (i.e., both directions are equally supported), leave the edge undirected.
- **Final PDAG Requirement:** The final PDAG must include exactly the edges provided in the causal skeleton, with no additional orientations beyond what is compelled by the evidence.

**Decision Flow:**
1. Identify and orient all edges that form the given v-structures.
2. Apply Meek rules to propagate any forced orientations.
3. For each remaining edge, determine if its orientation is uniquely dictated by the rules. If not, leave it undirected.
  
**Example:**  
If the edge A-B does not appear in any v-structure and both A -> B and B -> A would satisfy the constraints, do not orient it; keep it as A-B.

**Required Output Format:**
After completing your analysis, provide your final answer in this exact JSON format:
```json
{{
  "final_graph": {{
    "directed_edges": [
      {{
        "from": "Node1",
        "to": "Node2"
      }},
      {{
        "from": "Node2",
        "to": "Node3"
      }}
    ],
    "undirected_edges": [
      ["Node3", "Node4"],
      ["Node5", "Node6"]
    ]
  }}
}}
 ```
  
**Inputs:**
Premise: {premise}
(Contains the relevant marginal and conditional independence information.)
  
V-Structures: {v_structures}
(A list of collider patterns that must be preserved.)
  
Casual skeleton:
```json
{{
  "nodes": {nodes},
  "edges": {edges}
}}
```
\end{PromptListing}

\subsection*{Stage 4: Hypothesis Evaluation}

\begin{PromptListing}[caption={Hypothesis evaluation stage prompt template .},label={lst:prompt-hypothesis}]
You are a specialist in causal discovery algorithms, particularly the Peter-Clark (PC) algorithm, with a proven ability to evaluate complex causal relationships. In this final stage, your task is to assess a specific hypothesis regarding causal relationships between variables. Based on the constructed and oriented graph-derived from correlation statements, independence assertions, v-structures, and the application of Meek rules-determine whether the evidence supports or contradicts the proposed causal hypothesis.

**Task:** Evaluate whether the given causal hypothesis is supported by the provided causal graph.

**Context:**
You are given a causal graph that represents a Markov equivalence class (a CPDAG) derived from the above premises. This graph includes:
- **Directed edges:** These are relationships that are unambiguously oriented.
- **Undirected edges:** These represent ambiguous relationships where multiple orientations (across equivalent DAGs) are possible.
  
**IMPORTANT:** When evaluating the hypothesis, consider only those completions (i.e., fully oriented DAGs) that respect the above premises, including the separation sets. Do not allow any edge orientations that would violate these conditional independence statements (for example, avoid orienting the ambiguous edges as A -> B and C -> B simultaneously if that contradicts A _| C | B).
  
Your task is to determine if the specified hypothesis holds in every valid DAG within this equivalence class. The hypothesis is considered supported (true) only if it is true in all valid completions; if it holds in some but not all, then the answer should be false.
  
**Required Output Format:**
Provide your conclusion as a JSON object with a single boolean field:

```json
{{
  "hypothesis_answer": true/false
}}

**Inputs:**
Premise: {premise}
Provided Graph:
```json
{{
  "nodes": {nodes},
  "directed_edges": {directed_edges},
  "undirected_edges": {undirected_edges}
}}
```
Hypothesis: {hypothesis}
\end{PromptListing}

\section{Standard Deviation and Confidence Interval Report}
\label{appendix:stat_significance}

For the single‐prompt baseline experiment (Section \ref{subsec:baseline}), we report the bootstrap mean F1 for the DeepSeek-R1 API. Using five bootstrap samples ($R=5$, $B=1000$), we obtain a mean F1 of $0.6527$ (standard deviation $0.0102$) with a $95\%$ confidence interval of [$0.6461$, $0.6616$]. We believe that these results are representative of the remaining experiments in Table \ref{tab:results}; however, due to the high cost of additional runs and budget constraints, we only performed the full bootstrap analysis on the baseline experiment. In the final version, we will report the standard deviations for all the cases.

\end{document}